**Navigating the swarm: Deep neural networks command emergent behaviours**

**Authors:**
Dongjo Kim[1], Jeongsu Lee[2*], Ho-Young Kim[1,3*]

**Affiliations:**
[1] Department of Mechanical Engineering, Seoul National University, Seoul 08826, Korea
[2] Department of Mechanical, Smart, and Industrial Engineering, Gachon University, Seongnam 13120, Korea
[3] Institute of Advanced Machines and Design, Seoul National University, Seoul, 08826, Korea
* Corresponding authors. Email: leejeongsu@gachon.ac.kr; hyk@snu.ac.kr

**SUMMARY**

Interacting individuals in complex systems often give rise to coherent motion exhibiting coordinated global structures[1,2]. Such phenomena are ubiquitously observed in nature, from cell migration[3,4], bacterial swarms[5-7], animal and insect groups[8-23], and even human societies[24]. Primary mechanisms responsible for the emergence of collective behavior have been extensively identified, including local alignments based on average or relative velocity[5,12-17,25,26], non-local pairwise repulsive-attractive interactions such as distance-based potentials[27-29], interplay between local and non-local interactions[8,21-24,30-32], and cognitive-based inhomogeneous interactions[33,34]. However, discovering how to adapt these mechanisms to modulate emergent behaviours remains elusive. Here, we demonstrate that it is possible to generate coordinated structures in collective behavior at desired moments with intended global patterns by fine-tuning an inter-agent interaction rule. Our strategy employs deep neural networks, obeying the laws of dynamics, to find interaction rules that command desired collective structures, including rings, mills, clumps, and flocks. The decomposition of interaction rules into distancing and aligning forces, expressed by polynomial series, facilitates the training of neural networks to propose desired interaction models. Presented examples include altering the average radius and size of clusters in vortical swarms, timing of transitions from random to ordered states, and continuously shifting between typical modes of collective motions. This strategy can even be leveraged to superimpose collective modes, resulting in hitherto unexplored but highly practical hybrid collective patterns, such as protective security formations. Our findings reveal innovative strategies for creating and controlling collective motion, paving the way for new applications in robotic swarm operations[35,36], active matter organisation, and for the uncovering of obscure interaction rules in biological systems.

**TEXT**

Individuals in a group influence each other constantly. Collective motion emerges as a reflection of group-level consensus from individual interactions in the absence of centralized decision-making as seen in school of fish, flock of birds, and glow of fireflies (Fig. 1a). Elucidating the fundamental principles underlying the attainment of global consensus through the intricate network of individual interactions has long been an active area of inquiry in the study of collective motion[5,7-10,12-16,21-34,37-42]. Elementary interaction mechanisms that promote these collective behaviours include distancing in pairwise interactions based on relative positions[27-29] and aligning moving direction to an average value within an interaction range[5,12-17,25,26]. Despite our growing understanding of the fundamental principles governing the emergence of collective motion in complex systems, our ability to control these behaviours by adjusting individual-level interactions still falls short of our aspirations. The capability to actively manage these emergent behaviours has wide-ranging implications, including the development of decentralised control systems in robotic swarms, synchronization of facilities such as energy plants, and infrastructure management in mega-cities, such as traffic flow controls.

This study seeks a strategy to actively control collective behaviours of agents following physical rules by manipulating individual-level interaction rules. Our approach starts with a reciprocal interaction model that accounts for both distancing and aligning forces in agent interactions, represented as $\mathbf{F}_{ij} = f(r_{ij})\mathbf{e}_{\mathbf{r}_{ij}} + \sigma \circ g(r_{ij})\mathbf{e}_{\mathbf{v}_{ij}}$. These forces are based on the relative distance, $\mathbf{r}_{ij}$, and relative velocity, $\mathbf{v}_{ij}$, between interacting agents indexed as $i$ and $j$. Here, $\mathbf{F}_{ij}$ represents the pairwise interaction force, and $\mathbf{e}_{\mathbf{r}_{ij}}$ and $\mathbf{e}_{\mathbf{v}_{ij}}$ are unit vectors pointing in the directions of the relative distance and velocity, respectively. The functions $f$ and $g$ respectively describe the coefficients of distancing and aligning forces based on absolute distance $r_{ij} = |\mathbf{r}_{ij}|$. Additionally, $\sigma \circ$ signifies a cut-off operator that nullifies output beyond a specific interaction range $r_c$, ensuring that alignments only occur within physically realistic limits. In this description, our primary goal is to propose a strategy for determining the mapping functions $f$ and $g$ that govern the entire agent group to promote desired collective patterns.

Discovering the functional relationships for $f$ and $g$ requires consideration of both the physical laws that describe dynamic behaviours and the geometric parameters that characterize collective patterns. Physics-informed neural networks[43-45] enable the exploration of a solution curve that adheres to both physical rules and data-driven features by integrating these elements into the loss function. Inspired by this concept, we employed neural networks to express the positions ($x_i$ and $y_i$) of interacting agents over time $t$ for $i$-th agent. We then integrate the equation of motion (Eq. 1 in Methods) of interacting agents into the loss function, drawing on the self-propelled particles model[4,5,12-17,19,21,23-25,27,28,31,33], as depicted in Fig. 1b.

Next, the coefficient functions $f$ and $g$ are represented using a polynomial series: $f(r_{ij}) = \sum_k a_k r_{ij}^{n_k}$ and $g(r_{ij}) = \sum_k b_k r_{ij}^{m_k}$. These polynomial representations provide the flexibility needed to accurately model the random functional and facilitate neural network training. The coefficients, $a_k$ and $b_k$, along with the exponents of the polynomials, $n_k$ and $m_k$, become trainable parameters within the neural networks. Therefore, the neural networks in our study are expressed as $(x_i, y_i) = \text{NN}(t \mid a_k, b_k, n_k, m_k, \theta)$, where $\theta$ represents the trainable parameters in neural network layers. The detailed description is appended in Methods.

The neural networks are trained from random initial states to generate desired collective patterns at specified moments, as depicted in Fig. 1c. Simulations of self-propelled particle (SPP) dynamics are conducted to verify that the proposed interaction force model indeed leads to the emergence of collective motion. It is observed that a force model with randomly initialized $a_k$, $b_k$, $n_k$ and $m_k$ fails to produce coherent motion, leading to divergence among the agents. Training is accomplished by incorporating both geometric and dynamic characteristics of the target collective patterns into the loss function. The chosen target patterns are derived from canonical collective motions identified in previous studies[1,12-18,21,25-28,32,34,37], namely ordered state, ring, clumps, mill and flock as illustrated in Fig. 1d. The characteristic equations for the geometry and dynamics of each collective motion, which are integrated into the loss function, are summarized in Methods section and Extended Data Tables 1, 2.

Figure 2a illustrates the step-by-step procedure for modulating emergent behaviours. Initially, we select the target collective motion and then determine the corresponding force model, which may involve aligning, distancing, or a combination of these forces. The aligning force facilitates the emergence of a globally ordered state from a disordered group of active materials[1,25,26], whereas the distancing force promotes the formation of collective swarms, which are categorized as rings, clumps, and mills[27,28,32,37]. In addition, the flocking behaviour, characterized by the exploration of unbounded regions similar to formation flight, is induced by the combination of aligning and distancing forces. Subsequently, we incorporate the desired characteristics of the collective motion—such as the timing of the onset of the ordered state, rotational direction, and geometric features of the swarm—into the loss function of the neural networks. The neural networks then identify the force model best suited to achieve the desired collective pattern.

We first present the emergence of an ordered state from a completely disordered state, initiated by localized alignments in scenarios where $f = 0$, as depicted in Fig. 2b and supplementary video 1. Previous studies have demonstrated that localized aligning actions can induce a globally ordered state, as observed in ferromagnetic colloids[46], vibrating granules[47,48], and liquid crystals[49,50]. This experiment aims to show that our strategy can directly control the timing of the transition to a globally ordered state, defined as the time for order parameter $O = (1/N)|\sum_i v_i/v_i|$ to reach 0.9. We examined the movement of 40 SPPs within periodic boundaries as a model for phase transition[1,25]. Two distinct successful cases are presented: 1) controlling the timing to achieve a globally ordered state at a fixed interaction range, and 2) synchronizing the duration to reach the ordered state across cases with varying interaction ranges (Fig. 2c, d). Our strategy identifies an adjustable input parameter of timing for the neural networks, exhibiting a linear correlation with the time required to transition from a disordered to an ordered state (Fig. 2c, top). Moreover, we can fine-tune the timing for the appearance of the ordered state to achieve precision within 4.7% error with average value 17.8 across various interaction ranges (Fig. 2c, bottom), and confirmed the robustness of the trained interaction models' ability to modulate ordered state by conducting 100 trials under random initial condition, and noise; see Extended Data Fig. 1.

Next, the collective patterns that emerge from distancing forces in scenarios where $g = 0$ are examined, as detailed in Fig. 3a-c and supplementary video 2-6. These patterns correspond to ring, clump, and mill formations. The emergence of each collective pattern is characterized by a threshold distance where the attraction force transitions to repulsion force. For the ring pattern, only attraction forces are present, as indicated by the negative values in the $f(r_{ij})$ plots of Fig. 3a. The magnitude of the attraction force increases with the distance and decreases with the desired pattern radius. The clump and mill patterns emerge from a combination of attraction and repulsion in distancing forces, each with a distinct critical transition point (Fig. 3b, c). The relative scale of repulsion versus attraction is known to trigger the transition from clumps to mill patterns[28,37], corroborating the predictions of our proposed neural networks. These experiments confirm the efficacy of the proposed strategy in modulating emergent behaviours resulting from both aligning actions and distancing forces.

The proposed strategy can not only give rise to the desired collective pattern but also controls its geometric features. As shown in Fig. 3a-c, the global radius of swirling patterns, denoted as $R$, has been effectively manipulated to the desired size. The plot of the global radius versus the desired radius confirms a clear match for every collective pattern considered (ring, clumps, and mill). Furthermore, the local geometric and dynamic features can also be controlled, as demonstrated in Fig. 3b, c. For instance, the size of clusters $\varepsilon$ in the clumps can be independently controlled by incorporating the standard deviation of the radius distribution into the neural networks' control parameters (Fig. 3b). The neural networks have effectively demonstrated the ability to modulate the size of clusters by tuning the relative scale of attraction to repulsive forces, as illustrated in the $f(r_{ij})$ plots.

For the mill pattern, the rotational direction could serve as an additional desired input. The subtle changes in force model enable a single mill characterized by unidirectional rotation, as depicted in Fig. 3c. The motion of SPPs begins from initial random states, balanced in total momentum, and is conducted under the influence of pairwise interactions that conserve this total momentum. Therefore, it is remarkable that the proposed strategy guides force model to achieve biased rotational momentum, resulting from the internal acceleration and deceleration of each agent.

The combination of distancing and aligning forces induces a flocking pattern in agents moving in unbounded environments, as depicted in Fig. 3d and supplementary video 7. In our study, we distinguish 'flock' from an 'ordered state.' Flock refers to an animal swarm model that explores open environments with a coherent geometric feature, whereas the ordered state represents the collective pattern of moving particles in a localized area of active matter, modeled with periodic boundaries. An aligning force effectively directs the swarm towards the same direction, resulting in a flocking pattern. A significant finding of this study is the ability of neural networks to determine a control strategy for adjusting the extent of spatial spread. The networks use the transition point from attraction to repulsion to precisely control the geometric size of the entire swarm in space, as illustrated in the coefficient of force plots in Fig. 3d. This control is executed by managing the maximum spatial extent of the swarm relative to its centre of mass, which is integrated into the loss function of the neural networks. Repeated experiments starting from random initial conditions confirm that the extent of spatial spread closely aligns with the desired swarm size, as shown in Fig. 3d. These experiments have confirmed that the proposed strategy could accurately modulate not only the appearance but also the specific geometric features of desired collective patterns.

We extend our strategy to actively leverage collective patterns in realizing hitherto unexplored collective systems, as illustrated in exemplary cases of Fig. 4. The first scenario in Fig. 4a involves continuous transitions between different collective modes to demonstrate a model of robotic swarms performing various functions via collective behaviors. The simulation reveals that the SPPs continuously alter global patterns by adjusting interaction force models under the guidance of neural networks. Initially, a scenario describing the desired collective patterns at specific temporal moments are provided to the neural networks. The neural networks then prompt interaction models for the corresponding collective patterns, having been trained on the geometric characteristics of the desired collective motions. Observations of the order parameter, normalized angular momentum $O_r = (1/N)\sum_i(\boldsymbol{r}_i \times \boldsymbol{v}_i/r_i v_i)$, and normalized absolute angular momentum $O_{r,\text{abs}} = (1/N)\sum_i|\boldsymbol{r}_i \times \boldsymbol{v}_i/r_i v_i|$ confirm the emergence of the desired collective patterns at the corresponding moments, namely a continuous shift from an initial random state to clumps, rings, mills, flock, and finally to a single mill. This experiment suggests that our strategy can dynamically control swarming behaviours by tuning the interaction rules of interacting agents.

The second scenario, depicted in Fig. 4b, explores the superimposition of distinct collective patterns to synthesize novel hybrid collective dynamics. The SPPs are divided into two groups, with each group adhering to a different interaction model specific to their assigned collective pattern. This process leads to non-homogeneous interactions. Despite the non-homogeneous nature of the interaction mechanism, each SPP interacts with every other agent, irrespective of group classification. In other words, each individual agent decides its motion based on the information from every other recognizable agent, without regard to group classifications. Consequently, the interactions become non-reciprocal, leading to non-conservative interaction dynamics (refer to Eq. 5 in Methods section).

Several representative cases of superimposed collective patterns are displayed in Fig. 4b. We first synthesized a dual-ring (i) and dual-clumps (ii) mode characterized by identical collective patterns, yet with two distinct geometrical features. These structures, while having different radii, share an identical centre. Each ring or clumps rotates independently but together they form a globally coherent pattern. Moreover, our strategy facilitates the combination of different collective modes, as demonstrated by the coexistence of ring and mill patterns. These patterns are controlled to share the same centre but maintain different global radii, resulting in a coherent pattern where the ring encircles the milling group from the outside (vii). Such collective patterns can be utilized to enable swarming agents to function collaboratively, for instance, one group protecting another or encircling a target group. Additionally, the synthesized collective patterns can be used to navigate an open environment in specific hybrid formation. The combination of flock and mill patterns (iv, viii) demonstrates a reconnaissance formation, effectively guarding a critical group by encirclement with another group.

This study presents a unique strategy to precisely control the collective behaviour of complex systems. By integrating distancing and aligning forces into the interaction force model, we enable neural networks to identify individual-level interaction models that generate desired collective patterns. This proposed strategy allows for the modulation of not only the timing of pattern occurrence and collective modes but also geometric and dynamic features, such as radius, local cluster size, and rotational direction. Furthermore, the strategy's versatility extends to applications such as the smooth transition between different collective modes within the same agent group, demonstrating potential for controlling swarming robots. The synthesis of collective patterns, aimed at provoking unconventional collective behaviours, has been demonstrated—for instance, in search-and-rescue formations or in guarding or capturing scenarios. For the first time, the proposed strategy successfully controls collective behaviours by adjusting the individual-level interaction rules. These findings are thus expected to shift our perspective on the emergence of complex systems from merely subjects of study to objects of use and control.

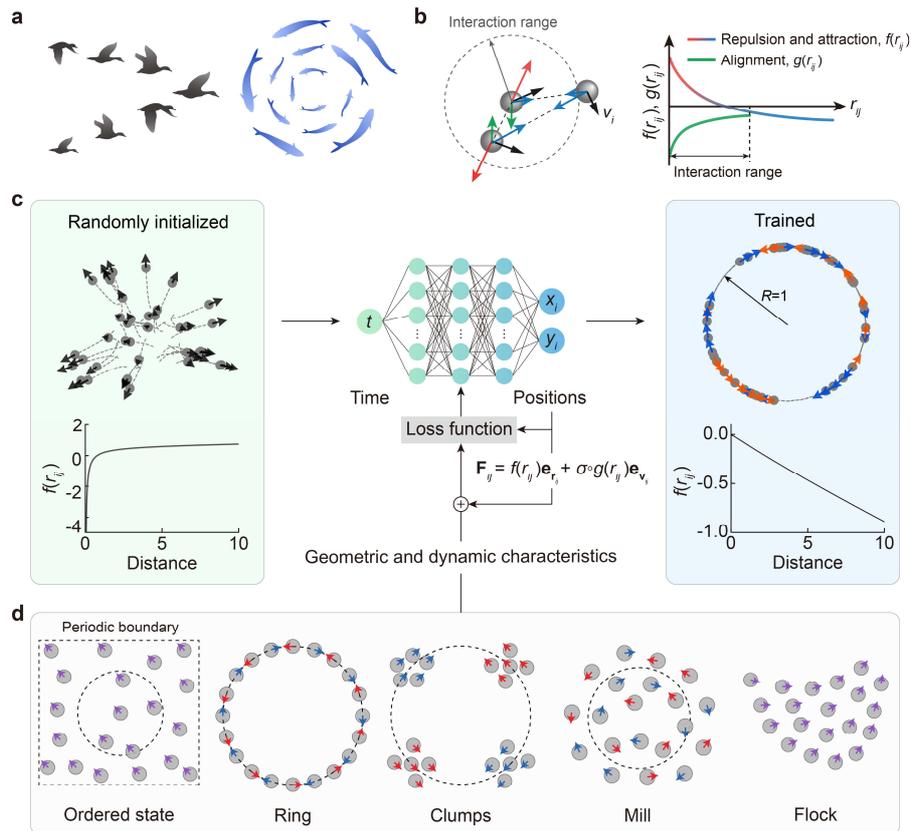

**Fig. 1| Schematic Illustration of the Strategy for Controlling Collective Motions. a,** Examples of collective behaviours, such as bird flocks and fish schools. **b,** Interaction model between agents, consisting of distancing and aligning forces. **c,** Procedure for constructing neural networks to realize desired collective patterns. This involves refining a randomly initialized interaction model through neural network training, incorporating the geometric and dynamic characteristics of the target collective patterns. **d**, Representative collective motions generated by the proposed strategy.

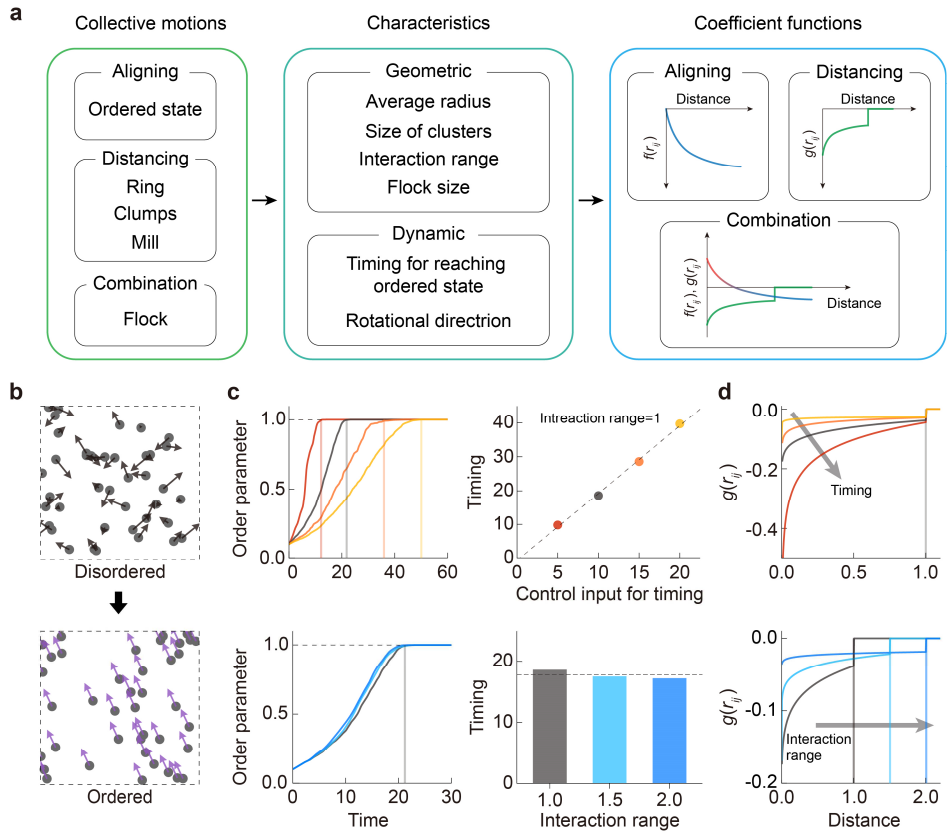

**Fig. 2| Modulation of emergent behaviour by manipulating interaction models. a**, Overall flowchart to obtain an interaction model to control collective behaviours. **b**, Illustration of the transition to an ordered state from an initially disordered state (number density $\rho = 4$); supplementary video 1. **c**, Order parameter plotted over time, manifesting the modulation of the timing for the appearance of the ordered state within a specified interaction range (top), and the synchronization of the emergence of the ordered state across different interaction ranges (bottom). **d**, Inferred interaction model for each collective motion in Fig. 2c.

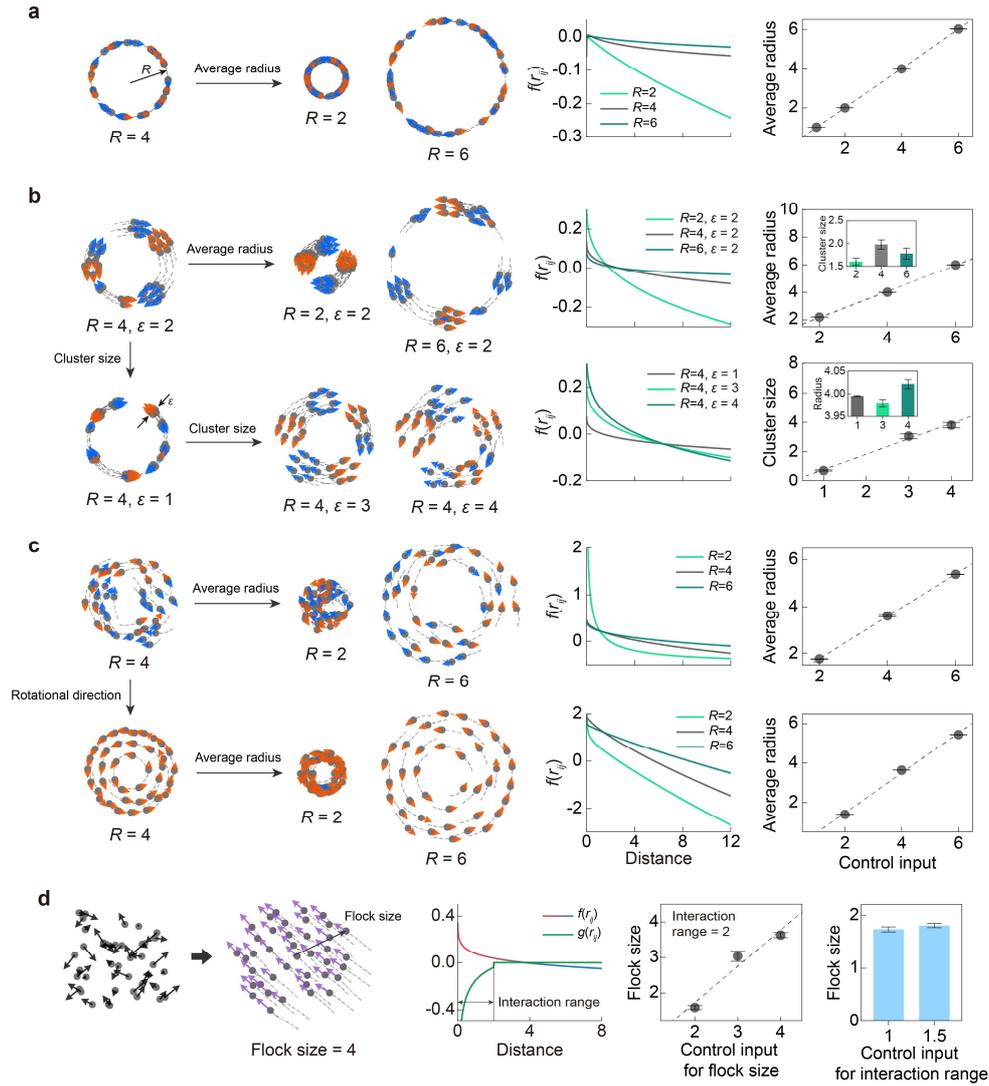

**Fig. 3| Visualization of collective behaviour patterns and their modulation.** Snapshots of ring, clump, mill, and flock patterns, as generated from numerical simulations and the interaction models identified via neural network training; supplementary video 2-6. Also, the average radius ($R$), the clusters' size for clumps ($\varepsilon$), and flock size (defined from the maximum spatial extent) are plotted against the control input for manipulation. The accompanying error bar on these plots signifies the standard deviation, derived from 100 trial runs. **a-c,** Observed collective motions. With a set radius $R = 4$, it is demonstrated how the geometric and dynamic properties of these patterns—namely rings, clumps, and mills—can be modulated to either decrease or increase in scale. **d**, The positions and velocities of 40 SPPs at $T = 0$ and 50 are graphically depicted, alongside the corresponding interaction models identified through neural network. All SPPs are randomly initialized at $T = 0$.

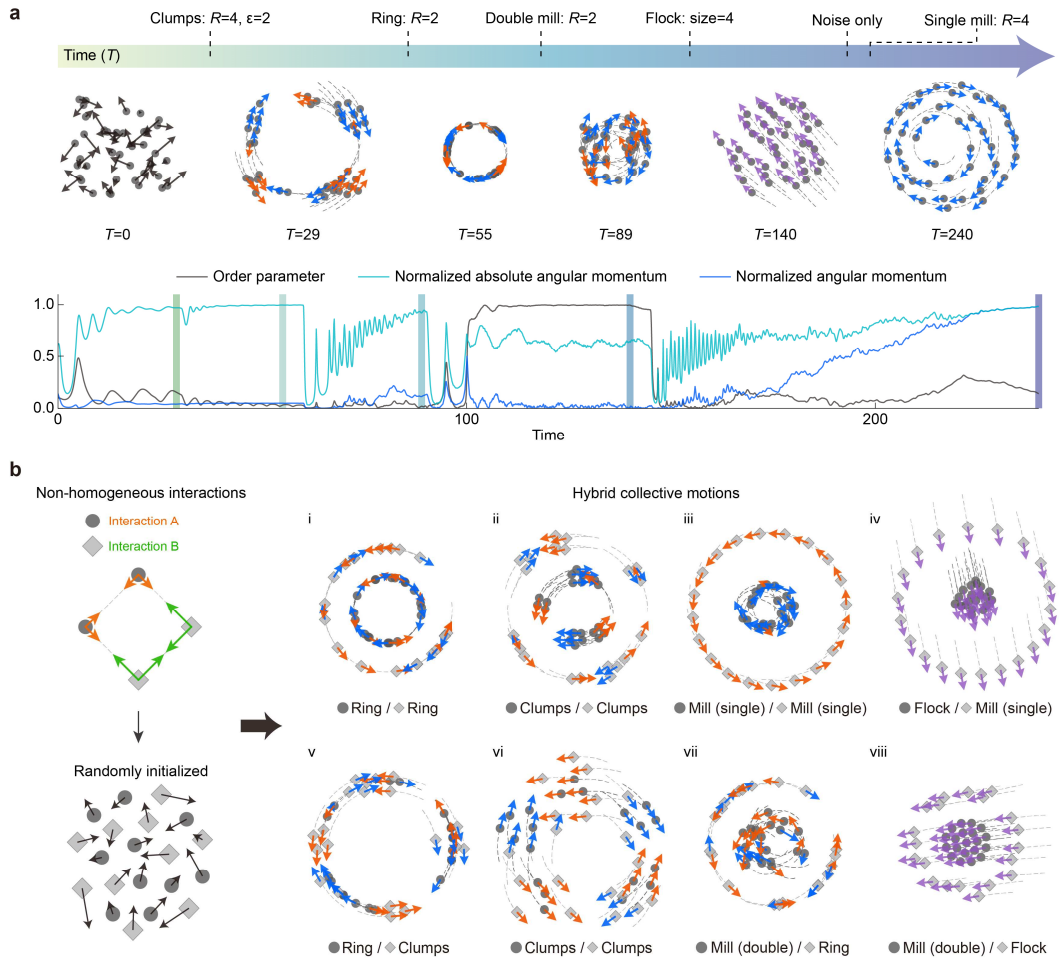

**Fig. 4| Versatile manipulations of collective motions. a**, Transition between collective motions triggered by changing the coefficients and exponents within the interaction model; supplementary video 8. It has been achieved by varying the coefficients and exponents in the interaction model over specified time intervals: clumps ($T = 0$-$30$), ring ($T = 30$-$60$), double mill ($T = 60$-$90$), flock ($T = 90$-$140$), and single mill ($T = 145$-$230$). The agents, initially in random positions and velocities, undergo these transitions smoothly. During a brief period of $T = 140$-$145$, where only noise is introduced to ensure unbiased conditions for the subsequent phase. Temporal variations in the order parameter, normalized absolute angular momentum, and normalized angular momentum signify the achievement of each target pattern when at least one of these parameter values approaches unity. **b**. Emergence of unprecedented hybrid collective patterns governed by non-homogeneous interaction models; supplementary video 9. Models A and B, developed through neural network training, are applied to two groups of 20 agents within a total of 40: ring ($R = 2$) – ring ($R = 4$), clumps ($R = 2$, $\varepsilon = 2$) – clumps ($R = 4$, $\varepsilon = 2$), single mill ($R = 2$) – single mill ($R = 4$), flock (size = 2) – single mill ($R = 4$), ring ($R = 4$) – clumps ($R = 4$, $\varepsilon = 2$), clumps ($R = 2$, $\varepsilon = 3$) – clumps ($R = 2$, $\varepsilon = 4$), double mill ($R = 2$) – ring ($R = 4$), and double mill ($R = 2$) – flock (size = 4).

**Acknowledgements**

This work was supported by National Research Foundation of Korea (Grant No. 2018-052541 and 2021-017476) via SNU SOFT Foundry Institute, and administered by SNU Institute of Engineering Research.

**Author contributions**
H.-Y.K. conceived the idea and supervised the research. D.K. and J.L. were responsible for designing the experiments and analyzing the data. D.K., J.L., and H.-Y.K. contributed to the writing of the manuscript.

**Code availability**
The Python code used for trainings and simulations in this manuscript is available at the following link: https://github.com/dongjokim03/Collective_motions_training.git

**Competing interests**
H.-Y.K. and D.K. are the inventors on the patent applications (#10-2024-0054253, Republic of Korea) submitted by SNU R&DB Foundation that covers neural network training method for collective motions and its applications.

## Methods

**Self-propelled particles**

Collective movements are predominantly seen in entities capable of generating momentum. Thus, previous studies have employed self-propelled particles as a model for simulating the agents exhibiting collective behaviours [4,5,12-17,19,21,23-25,27,28,31,33]. The self-propelled agents were assumed to be able to generate propulsion, whose concept can be represented in multiple forms, such as constant ($v = v_0$)[4,5,13,25], linear ($d\mathbf{v}/dt = [(v_0/v)-1]\cdot\mathbf{v}$)[1,17,24], linear with damping ($d\mathbf{v}/dt = \gamma[(v_0/v)-1]\cdot\mathbf{v}$)[33], and marginal control ($d\mathbf{v}/dt = [(v_0/v)-1]^n\cdot\mathbf{v}$, $n>1$)[17,28]. Recent research has shown that the concept of marginal control could be highly suitable for analyzing the observed correlations in bird flocking[17]. However, our study focuses on uncovering the interrelations among agents in collective behaviours, leading us to choose the simplest self-propulsion form: linear propulsion. For the velocity $\mathbf{v}_i$ of agent $i$, the reference self-propelled speed $v_0$, and the interaction forces $\mathbf{F}_{ij}$ between agents $i$ and $j$, their motions can be described by the ordinary differential equation similar to the Langevin equation[12,13,15,17]

$$\frac{d\mathbf{v}_i}{dt} = \mathbf{v}_i\left(\frac{v_0}{v_i} - 1\right) + \sum_{j(\neq i)} \mathbf{F}_{ij} + \boldsymbol{\eta}_i, \qquad (1)$$

where $\boldsymbol{\eta}_i$ is white Gaussian noise whose variance[17] is proportional to dimension $d$ and effective temperature $T$. In our study, we set the reference self-propelled speed $v_0$ as 2.

**Loss function for neural network**

Our study employs a simple neural network architecture comprising four fully connected layers, each with 32 neurons, designed specifically for inferring hidden interactions in collective motions. The first three layers feature tanh activation functions, while the final layer employs no activation function, instead incorporating a scale factor $R$, which corresponds to the target of geometric characteristics, such as the average radius and the size of flock. The network's total loss function integrates components from the governing dynamics in Eq. (1) and ground truth data relevant to characteristics of collective behaviour. This function is a summation of two parts: $\mathcal{L}_{total} = \mathcal{L}_{ground\ truth} + \mathcal{L}_{ODE}$. $\mathcal{L}_{ODE}$ represents the residual error between the predicted force and that calculated from agent dynamics:

$$\mathcal{L}_{ODE} = \frac{1}{NN_t}\sum_{j=1}^{N_t}\sum_{i=1}^{N}\left[\frac{d\mathbf{v}_i}{dt} - \mathbf{v}_i\left(\frac{v_0}{v_i} - 1\right) - \sum_{j(\neq i)}\mathbf{F}_{ij}\right]^2. \qquad (2)$$

$\mathcal{L}_{ground\ truth}$ consists of the initial condition, geometrical, and dynamic characteristics specific to the targeted collective motion. The training target changes depending on the collective pattern, leading to variations in $\mathcal{L}_{ground\ truth}$. This variation is formulated as,

$$\mathcal{L}_{ground\ truth} = \sum_{*}\alpha_*\mathcal{L}_*, \qquad (3)$$

where * indicates the index for the loss function defined by Extended Data Tables 1 and 2. The maximum radius and minimum distance in Extended Data Table 2 are applied using the rectified linear unit (ReLU) function applying thresholds $R_{max}$ and $d_{min}$. This is implemented as max(0, $r_i$ - $R_{max}$) for the maximum radius and max(0, $d_{min}$ - $r_{ij}$) for the minimum distance. In loss for order parameter in transient state, ground truth values are applied while linearly increasing from initial value to 1.

**Training of neural network**

The objective of the neural network is to predict collective behaviours within the time range $t \in [0, T]$. For this purpose, the time interval [0, T] is discretized into $N_t$ points and normalized to the range of [-1,1]. The trainable parameters, including $a_k$, $b_k$, $n_k$, and $m_k$, are uniformly random initialized within the range of -1 to 1 (ring, clumps, and mill) or -0.1 to 0.1 (ordered state and flock). The weights in $\theta$ are set using the Xavier uniform initialization method, while the biases in $\theta$ are initialized with a constant value of 0.1. The training process commences with the Adam optimizer, utilizing a learning rate of $10^{-3}$ for 200 epochs. Subsequently, the optimization method is switched to the L-BFGS algorithm, based on the Pytorch framework with tolerance $10^{-4}$ for $\mathscr{L}_{total}$. To determine collocation points, equidistant time interval $\Delta t = T/(0.1N_t)$ is selected, corresponding to 10% of the $N_t$ time steps without loss of initial and final index. Specifically, for training related to maximum radius and minimum distance in flock scenarios, these characteristics cannot be applied in transient motion. Therefore, collocation points are selected based on $\Delta t$ but within the range [$N_t/2$, $N_t$], accounting for 5% of all points.

**Numerical solution**

To examine the consequent collective behaviours of SPPs under the learned coefficient functions $f$ and $g$, we computed the numerical solutions for the dynamics of SPPs as dictated by Eq. 1. In this model, $\boldsymbol{\eta}_i$ represents white Gaussian noise, which is only applied in the ordered state and flock. This noise has a zero mean with a standard deviation of $\sigma = (2dT)^{1/2}$, where $d$ denotes the spatial dimension's degree and $T$ the effective temperature[17]. For the ordered state in our research, we utilized a standard deviation of $\sigma = 1$, and for flock behaviour, a standard deviation of 10 was employed. The numerical solutions were obtained using the Runge-Kutta 4th order method, as implemented in Scipy, with a maximum time step of 0.01. For patterns like ring, clumps, and mill, initial positions and velocities for the $x$ and $y$ axes were randomly assigned values within [-1, 1]. On the other hand, for ordered states and flocks, initial positions were chosen within [-L, L] to maintain a density of $\rho = N/L^2$, while velocities were selected from the range [-$v_0$, $v_0$].

**Probability for collective patterns**

In our study, agents influenced by the learned interaction forces generally displayed the targeted collective patterns with the specified geometrical or dynamical characteristics. However, random initial positions and velocities sometimes resulted in skewed linear or angular momentum, leading to unexpected collective behaviours, such as swarm-like patterns. To investigate this, we conducted a series of tests to assess the probability of achieving the intended collective patterns under random initial conditions, repeating these tests 100 times (as shown in Extended Data Fig. 2). Except for the case of single mill patterns at $R = 4$ and $R = 6$, the occurrence probabilities for all intended collective patterns were observed to be higher than 80%. In the test of single mill patterns, irrespective of the initial momentum imbalances, unintended collective patterns frequently emerged. Previous studies have indicated that interaction forces greatly exceeding the magnitude of self-propulsion can give rise to crystalline swarm structures[28]. Indeed, all the unexpected patterns observed at single mill with the target of $R = 4$ and $R = 6$ were swarm-like. Consequently, careful selection of the target average radius, which serves as an input for the neural network, is important for effectively controlling single mill patterns.

**Exponential transition**

The transition between different collective motions is achieved by altering the coefficients and exponents within the same polynomial form of interaction forces. To avoid abrupt switching while changing the coefficient functions $f$ and $g$, we utilized a simple exponential decay function. Specifically, for the $p$th collective motion commencing at time $t = t_p$, the interaction forces are modified as

$$\mathbf{F}_{ij,p} = \left[f_p(1 - e^{t_p-t}) + f_{p-1}e^{t_p-t}\right]\mathbf{e}_{\mathbf{r}_{ij}} + \sigma \circ \left[g_p(1 - e^{t_p-t}) + g_{p-1}e^{t_p-t}\right]\mathbf{e}_{\mathbf{v}_{ij}}. \quad (4)$$

Using the exponential decay approach, by the time $t$ reaches $t_p + 5$, the term $1 - e^{t_p-t}$ exceeds 0.99, effectively transitioning $\mathbf{F}_{ij,p}$ to represent the interaction forces for the $p$th collective pattern.

**Non-homogeneous interaction**

Numerous models have been proposed to explain collective behaviours in agents, yet most of them assume homogeneous interaction models, implying uniform function of force between agents[13-17, 24-28, 31, 32]. Building on the neural network learning of homogeneous interaction forces, we extended our approach to investigate unprecedented collective patterns featuring non-homogeneous interactions. In a system of $N = 40$ agents, we assigned different coefficient functions: those with subscript A to 20 agents for collective pattern A, and those with subscript B to the remaining 20 agents for pattern B, as detailed in Eq. (5).

$$\mathbf{F}_{ij} = \begin{cases} f_A \mathbf{e}_{\mathbf{r}_{ij}} + \sigma_A \circ g_A \mathbf{e}_{\mathbf{v}_{ij}}, & \text{for } 1 \leq i \leq 20, 1 \leq j \leq 40 \\ f_B \mathbf{e}_{\mathbf{r}_{ij}} + \sigma_B \circ g_B \mathbf{e}_{\mathbf{v}_{ij}}, & \text{for } 20 < i \leq 40, 1 \leq j \leq 40 \end{cases} \quad (5)$$

Our polynomial interaction model preserves the geometrical characteristics, even when the number of agents sharing the same interaction is halved compared to the training scenario. This not only demonstrates the potential of leveraging collective behaviour for novel patterns but also underscores the robustness and adaptability of our polynomial interaction model.

| * | $v_{i,\text{initial}}$ | $\frac{1}{N}\sum_{i=1}^{N}\mathbf{r}_i$ | $r_i$ | $\frac{1}{N}\sum_{i=1}^{N}r_i$ | $\sigma(r_i)$ | $r_{ij}$ | $\frac{1}{N}\sum_{i=1}^{N}\left|\frac{\mathbf{r}_i\times\mathbf{v}_i}{Rv_0}\right|$ | $\frac{1}{N}\left|\sum_{i=1}^{N}\frac{\mathbf{r}_i\times\mathbf{v}_i}{Rv_0}\right|$ |
|---|---|---|---|---|---|---|---|---|
| | | | | | | $f(r_{ij})=0$ | | |
| Ring | 1 | 0 | 5 | 0 | 0 | 0 | 5 | 0 |
| Clumps | 1 | 1 | 0 | 1 | 5 | 0 | 0 | 0 |
| Mill (double) | 1 | 1 | 0 | 1 | 0 | 5 | 5 | 0 |
| Mill (single) | 1 | 1 | 0 | 1 | 0 | 5 | 0 | 5 |

**Extended Data Table 1|** $\alpha_*$ of $\mathcal{L}_{\text{ground truth}}$ **for ring, clumps, and mill.**

| * | $\mathbf{r}_{i,\text{initial}}$ | $\mathbf{v}_{i,\text{initial}}$ | $\frac{1}{N}\sum_{i=1}^{N}\left|\frac{\mathbf{v}_i}{v_0}\right|$ | $\frac{1}{N}\sum_{i=1}^{N}\left|\frac{\mathbf{v}_{i,\text{final}}}{v_0}\right|$ | Max. radius | Min. distance |
|---|---|---|---|---|---|---|
| | | | $g(r_{ij}) \neq 0$ | | | |
| Ordered state | 5 | 5 | 5 | 5 | 0 | 0 |
| Flock | 1 | 1 | 5 | 0 | 5 | 5 |

**Extended Data Table 2|** $\alpha_*$ of $\mathcal{L}_{\text{ground truth}}$ **for ordered state and flock.**

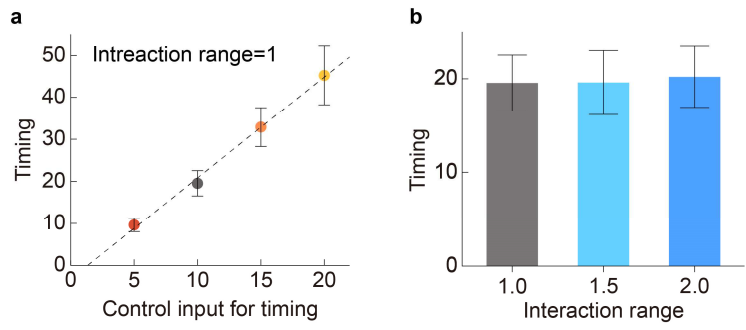

**Extended Data Fig. 1| Robustness of trained interaction model for ordered state. a**, Linear correlation between the control input and the timing for the ordered state. **b**, Timing versus interaction ranges. All error bars represent the standard deviation from 100 trials.

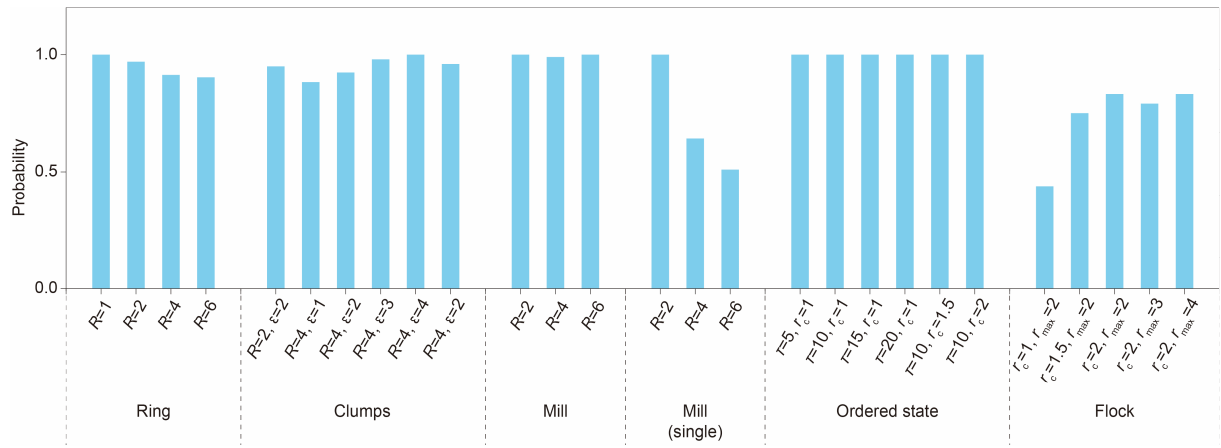

**Extended Data Fig. 2| Probability for collective pattern.** For 100 random initial conditions, 40 SPPs are simulated with trained coefficient functions *f* and *g*.